\documentclass[11
pt]{article}
\usepackage[utf8]{inputenc}
\usepackage[margin=1in]{geometry}
\usepackage{libertine}
\PassOptionsToPackage{numbers, compress}{natbib}
\usepackage{natbib}
\usepackage{hyperref}       
\usepackage{url}            
\usepackage{booktabs}       
\usepackage{amsfonts}       
\usepackage{nicefrac}       
\usepackage{graphicx}
\usepackage{subfigure}
\usepackage{amsmath}
\usepackage{amssymb}
\usepackage{multicol,multirow}
\usepackage[table]{xcolor}

\newcommand*{\affaddr}[1]{#1} 
\newcommand*{\affmark}[1][*]{\textsuperscript{#1}}

\title{Learning Clinical Concepts for\\ Predicting Risk of Progression to Severe COVID-19}
\author{%
Helen Zhou,\affmark[1] Cheng Cheng,\affmark[1] Kelly J. Shields,\affmark[2] \\
Gursimran Kochhar,\affmark[3] Tariq Cheema,\affmark[3] \\\vspace{0.5em}
Zachary C. Lipton,\affmark[1] Jeremy C. Weiss\affmark[1]\\
\affaddr{\affmark[1]Machine Learning Department, Heinz College, Carnegie Mellon University}\\
\affaddr{\affmark[2]Highmark Health Enterprise Data \& Analytics, Datasicence R\&D}\\
\affaddr{\affmark[3]Allegheny Health Network}\\
}
\date{ }

\begin{document}

\maketitle

\begin{abstract}
    With COVID-19 now pervasive, identification of high-risk individuals is crucial. Using data from a major healthcare provider in Southwestern Pennsylvania, we develop survival models predicting severe COVID-19 progression. In this endeavor, we face a tradeoff between more accurate models relying on many features and less accurate models relying on a few features aligned with clinician intuition. Complicating matters, many EHR features tend to be under-coded, degrading the accuracy of smaller models. In this study we develop two sets of high-performance risk scores: (i) an unconstrained model built from all available features; and (ii) a pipeline that learns a small set of clinical concepts before training a risk predictor. Learned concepts boost performance over the corresponding features (C-index 0.858 vs. 0.844) and demonstrate improvements over (i) when evaluated out-of-sample (subsequent time periods). Our models outperform previous works (C-index 0.844–0.872 vs. 0.598–0.810).
\end{abstract}

\section{Introduction}

As COVID-19 becomes endemic, communities are learning what it means for them to “live with” COVID-19. An important component of living with COVID-19 is understanding when individuals who contract the disease are likely to progress to a severe condition. Our work studies the risk of severe COVID-19 progression, using data collected by a major healthcare provider in Southwestern Pennsylvania from January 2020 to January 2022. We define \emph{severe COVID-19} as a COVID-19 case involving mechanical ventilation, admission to an intensive care unit (ICU), or death.

Healthcare providers often have different systems for collecting and storing patient data. To utilize this data for prediction,
researchers usually leverage domain expertise to manually extract a large initial set of potentially relevant features, and subsequently use automatic feature selection techniques to eliminate all but the most significant.
Clinicians can then 
consider these 
features when determining a patient’s care plan, and hospitals could potentially extract these features 
to calculate risk.
While expert-guided curation of features can help reduce the model search space, it can also limit performance due to imperfect feature extraction and inadvertent removal of informative features. Automatic feature selection may yield features that are predictive of the outcome, but these features may actually be serving as \emph{proxies} for higher-level concepts that cause the outcome (e.g. insulin medication may be predictive, but diabetes is the underlying risk factor), \emph{especially when the higher-level concepts are underreported.} While reliably recorded proxies can be effective predictors, they can also yield misleading interpretations. Additionally, it is unclear whether these proxies will be equally effective when applied to new settings such as different time periods or different hospitals. As a result, doctors may favor smaller models with features whose relevance is intuitive even if these models suffer some loss in performance owing, in part, to underreporting of the features.

To strike a balance between incorporating domain knowledge, model simplicity, transparency, and performance, we propose to learn clinical concepts anchored to intuitive expert-selected features, and to use these concepts to predict severe COVID-19 progression. Motivated by high levels of missingness in our data, clinical concepts are learned by treating the presence of an expert-selected feature (e.g. diabetes ICD code) as a positive label, treating its absence as unlabeled, and applying positive and unlabeled learning algorithms to learn the probability of the concept given the other covariates. We find that learned concepts (LC) for an expert-selected subset of features provide a boost in performance over the features (C-index 0.858 vs. 0.844), and that this boost places the LC model approximately halfway between the selected features model (C-index 0.844) and the model trained on all available features (All Features) or LC + All Features combined (both C-index 0.872). While there is some loss of performance going from the All Features model to the LC model, this gap seems to close quickly on subsequent time periods, suggesting that there may be some reason why the LC model is favored over time. Qualitatively, we find that some of the features important to the All Features model are incorporated into the learned concept classifiers, possibly indicating that they serve as proxies for the concepts. Finally, we publish an interactive web visualization tool at \url{acmilab.org/severe_covid} for users to explore the learned concepts, original features, and how both are utilized in our models. 

\section{Related Work}

Several works have identified predictive factors for severe COVID-19, where the population studied and the definition of severity vary. \citet{1docherty2020features} performed a prospective observational study on COVID-19 hospitalizations in the UK and identified risk factors for mortality including old age, male, and chronic comorbidities such as obesity. \citet{2henry2020hematologic} performed a meta-analysis of 21 studies and identified white blood cell count, lymphocytes, platelets, IL-6 and serum ferritin as inpatient biomarkers for progression to severe or fatal illness. The VACO Index\cite{3king2020development} uses three pre-COVID-19 health status variables, demographics, pre-existing medical conditions, and Charlson Comorbidity Index, in a mortality score. To identify severe COVID-19 patients in need of limited ventilation resources, some works\cite{4xu2021risk,5zhou2020mortality} have predicted patient risk of developing acute respiratory distress syndrome (ARDS) using labs, demographics, and other clinical data.

Covichem\cite{6bats2021covichem} is an admission risk score predicting severity as defined by a composite of lab values, ARDS, or ICU admission. After stepwise model selection on the Akaike Information Criterion, Covichem identified risk factors including: obesity, cardiovascular conditions, plasma sodium, albumin, ferritin, lactate, and creatinine. COVID-GRAM\cite{7liang2020development} predicts risk of ICU admission, invasive ventilation, or mortality for inpatients using ten predictors: chest radiography abnormality, age, hemoptysis, dyspnea, unconsciousness, comorbidity count, cancer history, neutrophil-to-lymphocyte ratio, lactate dehydrogenase, and direct bilirubin, chosen via LASSO regression. Galloway et. al.\cite{8galloway2020clinical} created a simple count-based risk score for predicting ICU admission or mortality, using twelve features: age, male, ethnicity, oxygen saturation, radiological severity score, neutrophils, C-reactive protein, albumin, creatinine, diabetes mellitus, hypertension, and chronic lung disease. Other works\cite{9salaffi2020role,10li2020clinical} have used chest CTs to score severity.

Several works have used deep learning to extract embeddings of medical concepts from EHRs.\cite{11choi2018mime,12rasmy2021med} While useful for various downstream tasks, these embeddings usually suffer a lack of transparency. As an alternative, Halpern et. al. proposed an “anchor-and-learn” framework in which expert-defined binary medical concepts are learned by treating certain informative features as positive labels for those concepts, and applying algorithms from positive and unlabeled learning.\cite{13elkan2008learning,14halpern2016electronic,15bekker2020learning} An advantage of this method is the interpretable coefficients of classifiers used for learning the concepts.

\newpage
\section{Data}

\paragraph{Cohort Description.}  We use retrospective observational data collected by a major healthcare provider in Southwestern Pennsylvania from January 1st, 2020 to January 12th, 2022. Out of 171,009 patients who were tested for COVID-19, we extract the 40,190 who tested positive. Of those, we remove individuals who were already mechanically ventilated or admitted to the ICU within 30 days prior to the time $t_0$ of testing positive for the first time. This leaves a cohort of 31,336 individuals (Table \ref{tab:cohort_characteristics}). Note that this study seeks to predict the risk of progressing to severe COVID-19 \emph{upon testing positive for the first time}, and so features and outcomes are defined relative to time $t_0$.

\paragraph{Features.}  Features are extracted no later than the date of each patient’s first covid positive test. These include testing location (inpatient/outpatient), demographics, labs, medications, vaccines, symptoms, and problem history. The most recent value of each feature is extracted, and symptoms are limited to a one-day window around $t_0$. Since there are tens of thousands of distinct medications, labs, diagnoses, vaccines, etc. in our data, the feature pool is limited to the top 20 of each data type except for labs (top 50). Upon clinician review, 45 more features are extracted. After removing low-variance features, converting categorical values to indicators, and normalizing continuous values, this yields a fixed-length 139-dimensional feature vector (see \url{acmilab.org/severe_covid}) for patient information known at $t_0$.

\paragraph{Outcome.} Since patients are right-censored upon leaving the hospital system, the outcome of interest is a time-to-event, where the time is computed as the time elapsed between $t_0$ and severe COVID-19 (mechanical ventilation, ICU admission, or death) or censorship (when the patient was last seen in hospital records), whichever is first.

\begin{table}[h]
\caption{Cohort characteristics (n = 31,336). Demographics, inpatient vs. outpatient status, outcomes.}
\vspace{0.5em}
    \label{tab:cohort_characteristics}
    \centering
    \begin{tabular}{lc}
    \toprule
Characteristic         & Count (\%)      \\
\midrule
\textbf{Gender}                 &                 \\
\hspace{1em}Female                 & 17,874 (57.0\%) \\
\hspace{1em}Male                   & 13,455 (42.9\%) \\
\textbf{Age}                    &                 \\
\hspace{1em}Under 20               & 2,836 (9.1\%)   \\
\hspace{1em}20 – 30                & 3,987 (12.7\%)  \\
\hspace{1em}30 – 40                & 4,134 (13.2\%)  \\
\hspace{1em}40 – 50                & 4,155 (13.3\%)  \\
\hspace{1em}50 – 60                & 5,444 (17.4\%)  \\
\hspace{1em}60 – 70                & 5,017 (16.0\%)  \\
\hspace{1em}70 or above            & 5,763 (18.4\%)  \\
\textbf{Location of Test}       &                 \\
\hspace{1em}Inpatient              & 13,246 (42.3\%) \\
\hspace{1em}Outpatient             & 15,868 (50.6\%) \\
\hspace{1em}Unknown                & 2,222 (7.1\%)   \\
\textbf{Outcomes}               &                 \\
\hspace{1em}Severe COVID-19        & 5,272 (16.8\%)  \\
\hspace{1em}ICU Admission          & 4,811 (15.4\%)  \\
\hspace{1em}Death                  & 1,554 (5.0\%)   \\
\hspace{1em}Mechanical ventilation & 1,096 (3.5\%)  \\
\bottomrule
\end{tabular}
\end{table}

\section{Learning Clinical Concepts }
Different types of data often provide partial information about higher-level concepts. For example, a saline IV bolus is typically administered inpatient, and is highly predictive of inpatient status even if inpatient status is unavailable. Certain labs could further confirm inpatient status. While one could methodically create rules for every concept of interest, it is difficult to do so comprehensively. As a result, learned models may end up using proxies that indirectly encode important risk factors (e.g. IV bolus encoding inpatient status), possibly leading to misinterpretation. Thus, we learn clinical concepts corresponding to major risk factors, and use these for downstream risk prediction.

\paragraph{PU Algorithm for Learning Concepts.}  To learn these concepts, we use the “anchor-and-learn” framework.\cite{14halpern2016electronic} For each concept of interest, we identify some key informative observations (“positive anchors”) relating to that concept. In this work, we only consider binary-valued concepts (present vs. not present). An observation is an anchor for a concept if it is conditionally independent of all other observations conditioned on the concept. When the presence of an anchor almost certainly implies the presence of the concept, this is known as a \emph{positive anchor}. 

Consider a patient with covariates $x\in\mathbb{R}^d$. Suppose we want to extract a concept $c$ with positive anchor $x_c \in\{0,1\}$ (e.g. extracting a diabetes concept with a diabetes diagnosis code as a positive anchor). Let $y_c\in \{0,1\}$ be the true binary label for whether concept $c$ is present. Note that in most observational health data, we observe the presence of a clinical condition, but not the absence of it. For example, when extracting the diabetes concept, we can be fairly confident that a patient marked as diabetic does indeed have diabetes, but patients unmarked do not necessarily \emph{not have} diabetes. Said differently, we have positive and unlabeled (PU) data rather than positive and negative data. Since only positive examples are labeled, $y_c=1$ is certain when $x_c=1$, but when $x_c=0$, then $y_c$ could be either 0 or 1. 

Thus, we leverage algorithms designed to learn from positive and unlabeled data, or “PU learning” algorithms. Let $x_{\bar{c}}$ refer to all covariates except for $x_c$. Since anchors are conditionally independent of all other observations conditioned on the concept, we have that 
$p(x_c | y_c = 1) = p(x_c | y_c = 1, x_{\bar{c}})$.
Now, consider $p(x_c=1 | x_{\bar{c}}  )$. We have that:
\begin{align*}
    p(x_c=1|x_{\bar{c}}  ) 
    &=p( x_c = 1 \land y_c=1| x_{\bar{c}}  )\\
    &=p(y_c=1|x_{\bar{c}} )p(x_c=1|y_c=1,x_{\bar{c}} )\\
    &=p(y_c=1|x_{\bar{c}} )p(x_c=1| y_c=1)\\
    \implies p(y_c=1|x_{\bar{c}} )&=p(x_c = 1 | x_{\bar{c}}  ) / \delta_c
\end{align*}
 			      
where $\delta_c= p(x_c=1 | y_c=1)$. The first equality follows from the fact that $y_c=1$ is certain when $x_c=1$, and the second equality follows from Bayes rule. In words, the expression indicates that \emph{true probability of the concept} being present is \emph{proportional to the probability of the positive anchor being present} by a factor of $\delta_c$. Thus, if we can train a PU classifier $g(x_{\bar{c}} ) = p(x_c=1 | x_{\bar{c}}  )$ that learns the probability that a positive anchor is present given the remaining covariates, we need only scale the probability by $\delta_c$ in order to get the probability of the underlying concept being present. As noted in Elkan and Noto,\cite{13elkan2008learning} for the set $P$ of positive labeled examples, one can construct an empirical estimate of the constant $\delta_c$ as  $\hat{\delta}_ c= \frac{1}{n} \sum_{x_{\bar{c}} \in P }g(x_{\bar{c}})$ , due to the observation that $g(x_{\bar{c}})= \delta_c$ for $x_{\bar{c}} \in P$:
\begin{align*}
    g(x_{\bar{c}} ) 
     &=p(x_c = 1 |  x_{\bar{c}}  )\\
 	&=p(x_c=1 | x_{\bar{c}} , y_c=1) p(y_c=1 | x_{\bar{c}}  )  + p(x_c=1 | x_{\bar{c}}, y_c = 0) p(y_c = 0 | x_{\bar{c}} )\\
 	&= p(x_c=1 | x_{\bar{c}}, y_c=1)\cdot 1 + 0 \cdot 0       \text{\hspace{1em} since $x_{\bar{c}} \in P$}\\
 	&=p(x_c=1 | y_c=1)\\
 	&=\delta_c.
\end{align*}
Finally, this yields the following procedure for learning clinical concepts:
\begin{enumerate}
	\item \textbf{Identify clinical concepts of interest}, and corresponding positive anchors.
	\item \textbf{Learn a positive vs. unlabeled classifier.} Use logistic regression to learn a classifier $g(x_{\bar{c}})$ that outputs the probability of the positive anchor given the other covariates.
	\item \textbf{Estimate the scaling constant.} On a validation set, estimate $\hat{\delta}_c$  by averaging the output of $g(x_{\bar{c}})$ on all positive labeled examples (i.e. examples with the positive anchor).
	\item \textbf{Scale predictions from the PU classifier} by the estimated scaling constant to get the probability that the underlying concept is present.  That is, compute $p(y_c=1 | x_{\bar{c}}) = g(x_{\bar{c}})/ \hat{\delta}_c$ for all examples where the positive anchor is not present. If the positive anchor is present, leave the probability as 1.
\end{enumerate}
This procedure is also used in Halpern et. al.,\cite{14halpern2016electronic} except instead of drawing the concept from a Bernoulli distribution parameterized by $p(y_c=1| x_{\bar{c}} $), we directly use the computed probability $p(y_c=1| x_{\bar{c}} )$ since it can provide more granular information. The scikit-learn\cite{16pedregosa2011scikit} python package was utilized for its logistic regression implementation.

\paragraph{Identifying Concepts of Interest.}
In order to define clinical concepts of interest, we surveyed several clinicians in the healthcare provider network about the main concepts they would look for when assessing risk of severe COVID-19. The survey yielded 21 concepts: old age, inpatient, outpatient, diabetes, shortness of breath, fever, cough, fatigue, COVID-19 vaccination, flu vaccination, obesity, hypertension, immunocompromised, COPD, congestive heart failure, chronic kidney disease, hyperglycemia, transplant, cancer, lung disease, and myalgia. We identify positive anchors for these concepts (precise definitions at \url{acmilab.org/severe_covid}) and apply the PU algorithm to extract a more complete representation of the concepts.
Learning Severe COVID-19 Risk
Using the lifelines python package,\cite{17davidson2019lifelines} a Cox proportional hazards model with L1 regularization (Lasso-Cox) is used to model risk of progression to severe COVID-19. For a patient with covariates $X$, their hazard $h$ at time $t$ is given by: 
$$h(t)=h_0 (t)  \exp(X\beta)$$
where $h_0$ is a baseline hazard function, and $\beta$ are learned coefficients. The regularization penalty is given by $\lambda ||\beta||_1$, where regularization strength $\lambda$ is selected using 5-fold cross validation and grid search over penalties between 0 and 0.2, with a step size of 0.001 between each penalty. For stability of training, features with variance $< 0.01$ are removed.

\section{Experimental Setup }

\paragraph{Feature Sets.} In order to explore the marginal effect of incorporating learned concepts versus the original set of 139 features, we analyze and evaluate Lasso-Cox models learned from five different sets of features:

\begin{enumerate}
    \item \textbf{Raw positive anchors}: only the positive anchors identified in the data, without learning the corresponding clinical concepts (e.g. mention of "diabetes" in a note, ICD code for diabetes, etc.)
	\item \textbf{Learned concepts (LC)}: only the learned clinical concepts (e.g. the diabetes concept)
	\item \textbf{LC + Numeric}: the learned clinical concepts and  numerical features (e.g. diabetes concept and labs)
	\item \textbf{LC + All Features}: the learned concepts, as well as all of the original 139 features
	\item \textbf{All Features}: all 139 original features, no learned concepts. 
\end{enumerate}

\paragraph{Back-testing and Data Splits.} In real-world settings, hospital systems may want to use updated data to revise their models. To emulate this process, we re-train models (including PU concept classifiers) up to the end of each 3-month season (spring, summer, fall, winter), and evaluate their performance on subsequent seasons. Spring is March 20th until June 21st, followed by summer until September 22nd, followed by fall until December 21st, followed by winter until March 20th of the following year. For each 3-month period, a 70-30 split designates train and test sets, where test data is never included in any model training. To keep the risk score interpretation simple, for each time period a grid search on the Lasso-Cox penalty is done to choose a model with approximately ten features. We additionally train models on the entire study time range, with train and test sets that aggregate the respective 3-month datasets.

\section{Evaluation}

\paragraph{Clinical Concept Evaluation.} Since the concepts are only positively labeled or unlabeled, it is not possible to compute precision of the concept classifiers.\cite{15bekker2020learning} However, we can compute recall as the proportion of known positives recovered by the classifiers. Additionally, we examine the number of previously unlabeled samples predicted to be positive.

\paragraph{Model Interpretation.} The Lasso-Cox model coefficients are in terms of original features as well as learned concepts. In addition to listing the Lasso-Cox coefficients, we create an interactive Sankey diagram to visualize how raw features translate into concepts, and how the resulting models pull from both. This gives the user a birds-eye view of how each concept is defined, the strength and sign of the coefficients, and which concepts are used in different models. 

\paragraph{Survival Model Evaluation Metrics.} The concordance, or C-index, is used to evaluate the model's discriminative ability. To evaluate calibration, both one-calibration at 14 days and D-calibration are used.\cite{18haider2020effective} Additionally, low, medium, and high-risk strata are defined and their 14-day Kaplan-Meier survival curves are inspected.

\paragraph{Baselines.} We compare our model performance to that of the Covichem\cite{6bats2021covichem} and Galloway\cite{8galloway2020clinical} risk scores. For Covichem, in order to conduct a fairer comparison than directly applying their logistic regression coefficients learned on a different population, we extract the same features and re-train logistic regression on our own training data. For Galloway, we extract all but one of their twelve features (radiological severity is not available in our data), and compare performance against two versions of their model: (1) directly applying their proposed count-based risk score (Galloway count), and (2) re-training a logistic regression model using the twelve variables (Galloway reweighted). 

\section{Results}
The PU learning algorithm yields concepts ranging from those with high recall, e.g. 0.974 for inpatient status, to low recall, e.g., 0.381 for immunocompromised (Table \ref{tab:num_pu_pos}). Some concepts have substantially more new positives (obesity, with 2,157 new positives). Concept classifier coefficients are available at \url{acmilab.org/severe_covid}. 

The model trained on learned concepts (LC) achieves a higher aggregate concordance than the original features corresponding to those concepts (0.858 vs. 0.844, Table \ref{tab:model_perf}). The model learned from all original features (All Features) and LCs (C-index 0.872) performs comparably to All Features alone (C-index 0.872). The addition of numerical features to the LCs does not significantly improve performance (C-index of both are 0.858). In all models except for the ones trained on All Features or All Features + LCs, the aggregate C-index is higher than the C-indices on the inpatient and outpatient subpopulations.

In the models using all features, medications such as dexamethasone, acetaminophen, and intravenous saline are selected (Table \ref{tab:hazard_ratios}). Across all models, the inpatient status is the feature with the greatest hazard ratio. Blood urea nitrogen is used by both the LC + All Features and All Features models.  Figure 1 is a screenshot of an interactive Sankey diagram which allows users to explore the coefficients for both the underlying clinical concepts and the classifiers built on top of the features and concepts. The interactive web tool is available at \url{acmilab.org/severe_covid}. 

When evaluated over time, the models with learned concepts achieve higher concordance several months after the model was initially trained, whereas the All Features model achieves higher concordance in the immediate term. For example, in Spring 2020, the concordance of the All Features model trained up until the end of Spring 2020 is 0.842, compared to 0.797 in the LC only model. By Fall and Winter 2021, however, the All Features model degrades to 0.808 or stays around 0.845, whereas the LC only model actually increases concordance to 0.83 and 0.904. Reading the table from left to right, the performance of any model fluctuates no more than 0.121 over all seasons. Reading the table from top to bottom, several columns shown an increase in performance as models are trained on more recent data.

The Kaplan-Meier curves corresponding to the low, medium, and high risk groups derived from the LC + All Features model and LC only model predictions are shown in Figure 2. There is a clear separation between the survival trajectories of the different risk groups. The LC + all features model appears to slightly under-estimate the risk of the high-risk groups, whereas the LC only model appears to be better calibrated at 14 days (Figure 3). The LC only model also appears to have better d-calibration across all time points (Figure 4).

\begin{table}[]
    \centering
    \caption{The number of new positives extracted by PU learning in the test set, the number of clinical concepts originally in the test set (determined solely by the presence of positive anchors), and the recall of the PU classifier among known positives in the test set. Concepts with prevalence $< 1.5\%$ are omitted.}
    \vspace{0.5em}
    \label{tab:num_pu_pos}
    \setlength{\tabcolsep}{12pt}
    \begin{tabular}{lccc}
    \toprule
         \textbf{Learned Concept}        & \textbf{New Positives}  & \textbf{Original Positives} & \textbf{Recall Among Original} \\
         \textbf{(LC)}                       & \textbf{(\% of Test Set)} & \textbf{(\% of Test Set)} & \textbf{Positives (count)} \\
         \midrule
        Old   age              & 147   (1.6\%)                  & 3,158   (33.7\%)                    & 0.956   (3,019)                         \\
        Inpatient              & 227   (2.4\%)                  & 3,914   (41.8\%)                    & 0.974   (3,813)                         \\
        Outpatient             & 207   (2.2\%)                  & 4,811   (51.3\%)                    & 0.961   (4,623)                         \\
        Diabetes               & 594   (6.3\%)                  & 834   (8.9\%)                       & 0.553   (461)                           \\
        Fever                  & 4,171   (44.5\%)               & 1,021   (10.9\%)                    & 0.826   (843)                           \\
        Shortness   of breath  & 1,518   (16.2\%)               & 1,005   (10.7\%)                    & 0.767   (771)                           \\
        COVID-19   vaccination & 2,128   (22.7\%)               & 1,884   (20.1\%)                    & 0.739   (1,392)                         \\
        Flu   vaccine          & 2,326   (24.8\%)               & 4,191   (44.7\%)                    & 0.864   (3,619)                         \\
        Obesity                & 2,157   (23.0\%)               & 433   (4.6\%)                       & 0.610   (264)                           \\
        Immunocompromised      & 909   (9.7\%)                  & 168   (1.8\%)                       & 0.381   (64)                            \\
        COPD                   & 575   (6.1\%)                  & 220   (2.3\%)                       & 0.623   (137)                           \\
        Hyperglycemia          & 496   (5.3\%)                  & 171   (1.8\%)                       & 0.737   (126)                           \\
        Cough                  & 4,023   (42.9\%)               & 1,862   (19.9\%)                    & 0.815   (1,517)                         \\
        Fatigue                & 2,947   (31.4\%)               & 602   (6.4\%)                       & 0.694   (418)                          \\
        \bottomrule
    \end{tabular}
\end{table}

\begin{table}[]
    \centering
    \caption{Performance of our models and baselines. The median C-index and 95\% CI are reported from bootstrapping the test set with 1000 replicates. Bold highlights the two models with highest C-index.}
    \vspace{0.5em}
    \setlength{\tabcolsep}{10pt}
    \label{tab:model_perf}
    \begin{tabular}{lccc}
    \toprule
         \textbf{Model}                      & \textbf{Aggregate   Test } & \textbf{Inpatient   Test } & \textbf{Outpatient   Test} \\
         & \textbf{C-index} & \textbf{C-index} & \textbf{C-index} \\
         \midrule
Covichem                   & 0.598 (0.580 – 0.616)    & 0.584 (0.569 – 0.600)    & 0.546 (0.509 – 0.581)     \\
Galloway count             & 0.745 (0.734 – 0.757)    & 0.647 (0.633 – 0.662)    & 0.714 (0.677 – 0.750)     \\
Galloway   reweighted      & 0.810   (0.803 – 0.824)  & 0.699   (0.673–0.703)    & 0.764   (0.728–0.709)     \\
Raw positive anchors       & 0.844 (0.836 – 0.851)    & 0.665 (0.650 – 0.680)    & 0.756 (0.709 – 0.796)     \\
Learned concepts (LC) only & 0.858 (0.851 – 0.865)    & 0.699 (0.685 – 0.713)    & 0.798 (0.757 – 0.834)     \\
LC + numerical features    & 0.858 (0.851 – 0.865)    & 0.695 (0.681 – 0.71)     & 0.814 (0.777 – 0.849)     \\
LC + all features          & 0.872 (0.865 – 0.877)    & 0.715 (0.702 – 0.728)    & 0.879 (0.858 – 0.901)     \\
All features (no LC)       & 0.872 (0.866 – 0.878)    & 0.717 (0.703 – 0.730)    & 0.880 (0.860 – 0.901)\\
\bottomrule
    \end{tabular}
\end{table}

\begin{table}[]
    \centering
    \caption{Hazard ratios (HR) of LC + All Features, LC, and All Features models. Abbreviations: Med = Medication, loc. = location, Dex. = Dexamethasone sodium phosphate, APAP = acetaminophen, SOB = Shortness of breath, BUN = blood urea nitrogen, NEUT = neutrophils, Immunocomp. = immunocompromised, vax = vaccine, OP = outpatient.}
    \vspace{0.5em}
    \label{tab:hazard_ratios}
    \resizebox{\columnwidth}{!}{%
    \begin{tabular}{|lc|lc|lc|}
\hline
\multicolumn{2}{|c|}{\textbf{All Features + LCs}}                                                                                      & \multicolumn{2}{c|}{\textbf{Learned Concepts (LC) Only} }                                                                 & \multicolumn{2}{c|}{\textbf{All Features Only}}                                                                                       \\ \hline
\multicolumn{1}{|l|}{Features}                                                                         & HR (95\% CI)         & \multicolumn{1}{l|}{Features}                                                             & HR (95\% CI)         & \multicolumn{1}{l|}{Features}                                                                         & HR (95\% CI)         \\ \hline
\multicolumn{1}{|l|}{(LC)   Inpatient}                                                                 & 2.31   (2.09 – 2.56) & \multicolumn{1}{l|}{(LC)   Inpatient}                                                     & 7.23   (5.43 – 9.62) & \multicolumn{1}{l|}{\begin{tabular}[c]{@{}l@{}}(Test   Location) \\ Inpatient\end{tabular}}           & 3.62   (3.31 – 3.97) \\ \hline
\multicolumn{1}{|l|}{(LC)   SOB}                                                                       & 1.72   (1.58 – 1.88) & \multicolumn{1}{l|}{(LC)   Old age}                                                       & 2.54   (2.33 – 2.77) & \multicolumn{1}{l|}{\begin{tabular}[c]{@{}l@{}}(Med)   Dex. \\ 4mg/mL \\ injection sol.\end{tabular}} & 1.91   (1.77 – 2.06) \\ \hline
\multicolumn{1}{|l|}{\begin{tabular}[c]{@{}l@{}}(Med)   Dex. \\ 4mg/mL \\ injection sol.\end{tabular}} & 1.54   (1.42 – 1.68) & \multicolumn{1}{l|}{(LC)   SOB}                                                           & 2.31   (2.14 – 2.49) & \multicolumn{1}{l|}{\begin{tabular}[c]{@{}l@{}}(Med)   APAP \\ 325mg tablet\end{tabular}}             & 1.67   (1.51 – 1.85) \\ \hline
\multicolumn{1}{|l|}{\begin{tabular}[c]{@{}l@{}}(Med)   APAP \\ 325 mg tablet\end{tabular}}            & 1.47   (1.34 – 1.62) & \multicolumn{1}{l|}{(LC)   Diabetes}                                                      & 1.28   (1.16 – 1.41) & \multicolumn{1}{l|}{Age   70+}                                                                        & 1.60   (1.49 – 1.72) \\ \hline
\multicolumn{1}{|l|}{(LC)   Old age}                                                                   & 1.34   (1.25 – 1.44) & \multicolumn{1}{l|}{(LC)   COPD}                                                          & 1.22   (1.13 – 1.32) & \multicolumn{1}{l|}{\begin{tabular}[c]{@{}l@{}}(Med)   NaCl \\ 0.9\% IV sol.\end{tabular}}            & 1.35   (1.24 – 1.46) \\ \hline
\multicolumn{1}{|l|}{\begin{tabular}[c]{@{}l@{}}(Med)   NaCl \\ 0.9 \% IV sol.\end{tabular}}           & 1.33   (1.23 – 1.44) & \multicolumn{1}{l|}{(LC)   Obesity}                                                       & 1.22   (1.14 – 1.3)  & \multicolumn{1}{l|}{(OP   ICD) SOB}                                                                   & 1.10   (1.01 – 1.19) \\ \hline
\multicolumn{1}{|l|}{(Lab)   BUN}                                                                      & 1.05   (1.01 – 1.1)  & \multicolumn{1}{l|}{\begin{tabular}[c]{@{}l@{}}(LC)   Immuno-\\ compromised\end{tabular}} & 1.10   (1.04 – 1.16) & \multicolumn{1}{l|}{(Lab)   BUN}                                                                      & 1.10   (1.05 – 1.14) \\ \hline
\multicolumn{2}{|l|}{\cellcolor[HTML]{D9D9D9}}                                                                                & \multicolumn{1}{l|}{(LC)   Fatigue}                                                       & 1.08   (1.02 – 1.15) & \multicolumn{1}{l|}{\begin{tabular}[c]{@{}l@{}}(Med)   Pantopra-\\ zole 40 mg tablet\end{tabular}}    & 1.05   (0.97 – 1.12) \\ \cline{3-6} 
\multicolumn{2}{|l|}{\cellcolor[HTML]{D9D9D9}}                                                                                & \multicolumn{1}{l|}{(LC)   Outpatient}                                                    & 1.06   (0.80 – 1.42) & \multicolumn{1}{l|}{\begin{tabular}[c]{@{}l@{}}(Lab)   NEUT \\ relative \%\end{tabular}}              & 1.04   (0.99 – 1.09) \\ \cline{3-6} 
\multicolumn{2}{|l|}{\cellcolor[HTML]{D9D9D9}}                                                                                & \multicolumn{1}{l|}{\begin{tabular}[c]{@{}l@{}}(LC)   Hyper-\\ glycemia\end{tabular}}     & 0.90   (0.81 – 1.00) & \multicolumn{1}{l|}{\begin{tabular}[c]{@{}l@{}}(Med)   NaCl \\ 0.9\% IV Bolus\end{tabular}}           & 1.04   (0.96 – 1.13) \\ \cline{3-6} 
\multicolumn{2}{|l|}{\cellcolor[HTML]{D9D9D9}}                                                                                & \multicolumn{1}{l|}{\begin{tabular}[c]{@{}l@{}}(LC)   COVID-19 \\ vax\end{tabular}}       & 0.87   (0.78 – 0.97) & \multicolumn{1}{l|}{(Lab)   Albumin}                                                                  & 0.98   (0.93 – 1.02) \\ \cline{3-6} 
\multicolumn{2}{|l|}{\cellcolor[HTML]{D9D9D9}}                                                                                & \multicolumn{1}{l|}{(LC)   Fever}                                                         & 0.82   (0.75 – 0.90) & \multicolumn{2}{l|}{\cellcolor[HTML]{D9D9D9}}                                                                                \\ \cline{3-4}
\multicolumn{2}{|l|}{\cellcolor[HTML]{D9D9D9}}                                                                                & \multicolumn{1}{l|}{(LC)   Flu vax}                                                       & 0.74   (0.66 – 0.83) & \multicolumn{2}{l|}{\cellcolor[HTML]{D9D9D9}}                                                                                \\ \cline{3-4}
\multicolumn{2}{|l|}{\multirow{-7}{*}{\cellcolor[HTML]{D9D9D9}}}                                                              & \multicolumn{1}{l|}{(LC)   Cough}                                                         & 0.58   (0.53 – 0.65) & \multicolumn{2}{l|}{\multirow{-3}{*}{\cellcolor[HTML]{D9D9D9}}}                                                              \\ \hline
\end{tabular}%
}
\end{table}

\begin{figure}
    \centering
    \includegraphics[width=\textwidth]{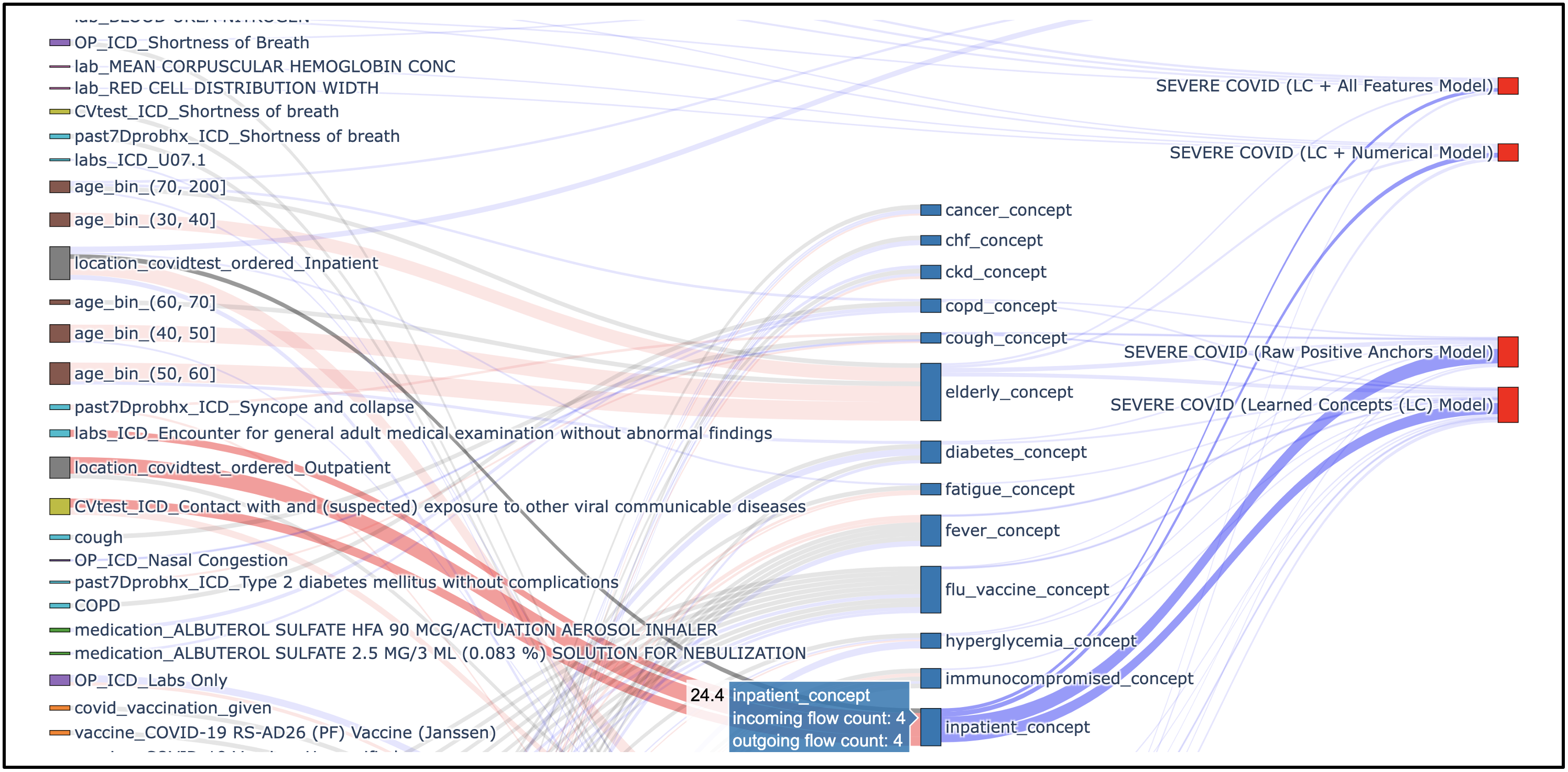}
    \caption{Screenshot of interactive Sankey diagram showing how raw features (first column) translate into clinical concepts (second column), and how both are ultimately used in each model (third column). Magnitude of coefficients correspond to flow thickness, positive log HRs are blue, and negative log HRs are red. Black flows indicate positive anchors for the corresponding concept. Visit \url{acmilab.org/severe_covid} to interact with the full diagram.}
    \label{fig:sankey}
\end{figure}

\begin{figure}
    \centering
    \includegraphics[width=\textwidth]{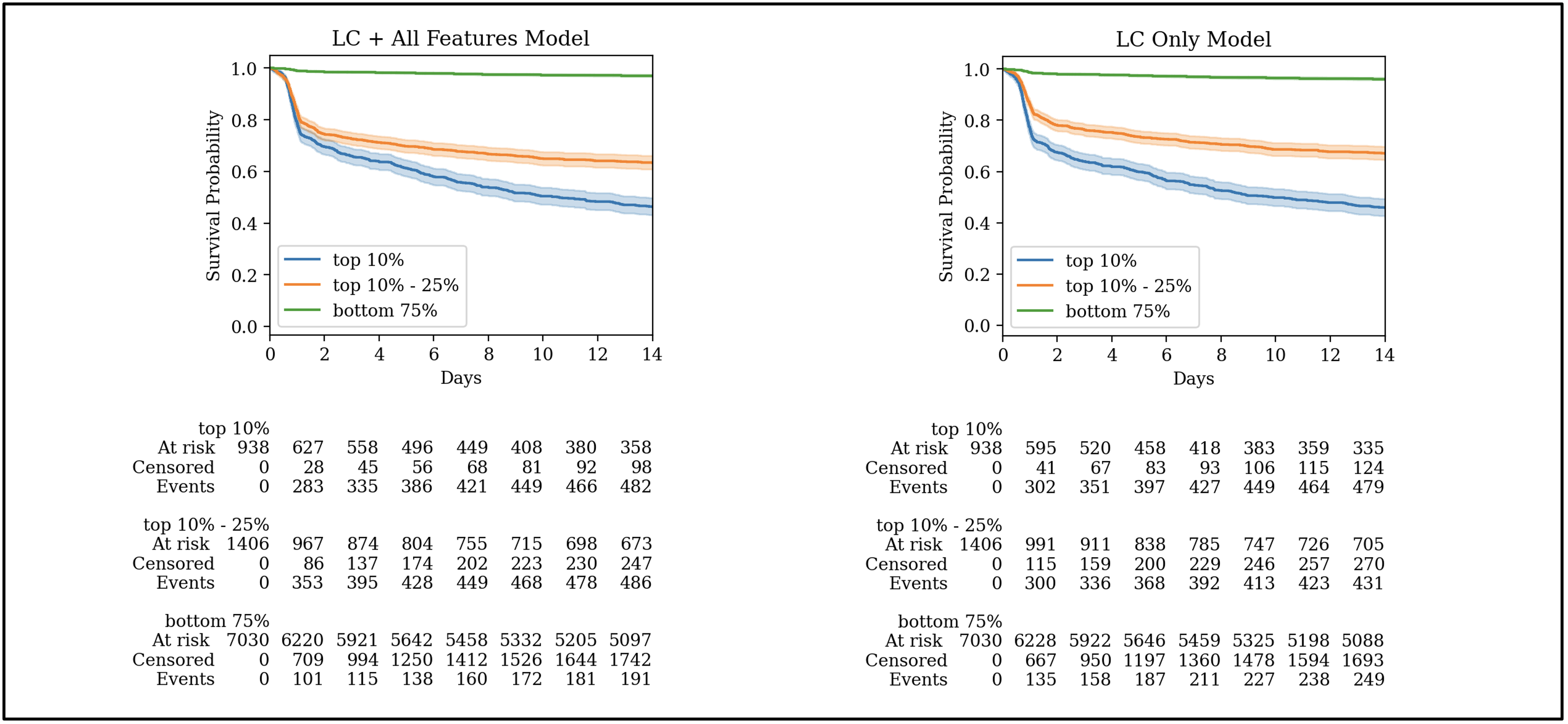}
    \caption{Kaplan Meier survival curves for the high (top 10\%), medium (top 10-25\%), and low (bottom 75\%) risk groups using predictions from the LC + All Features model (left) and the LC only model (right). Counts at the bottom show the number of individuals who are at risk, are censored, or experienced the severe COVID-19 event across time.}
    \label{fig:km_curves}
\end{figure}

\begin{figure}
    \centering
    \includegraphics[width=0.65\textwidth]{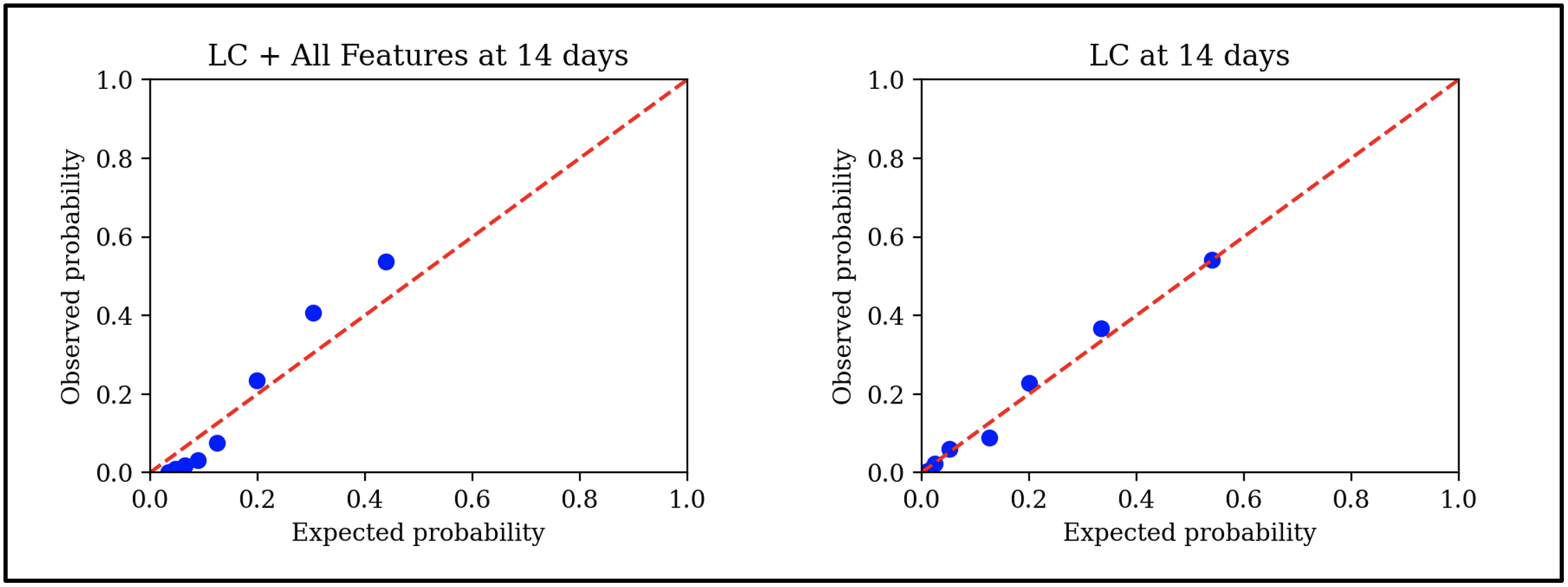}
    \caption{One-calibration of the LC + All Features model (left) and LC model (right) at 14 days, binned into ten groups. Red dotted line corresponds to perfect calibration.}
    \label{fig:calibration}
\end{figure}

\begin{figure}
    \centering
    \includegraphics[width=0.9\textwidth]{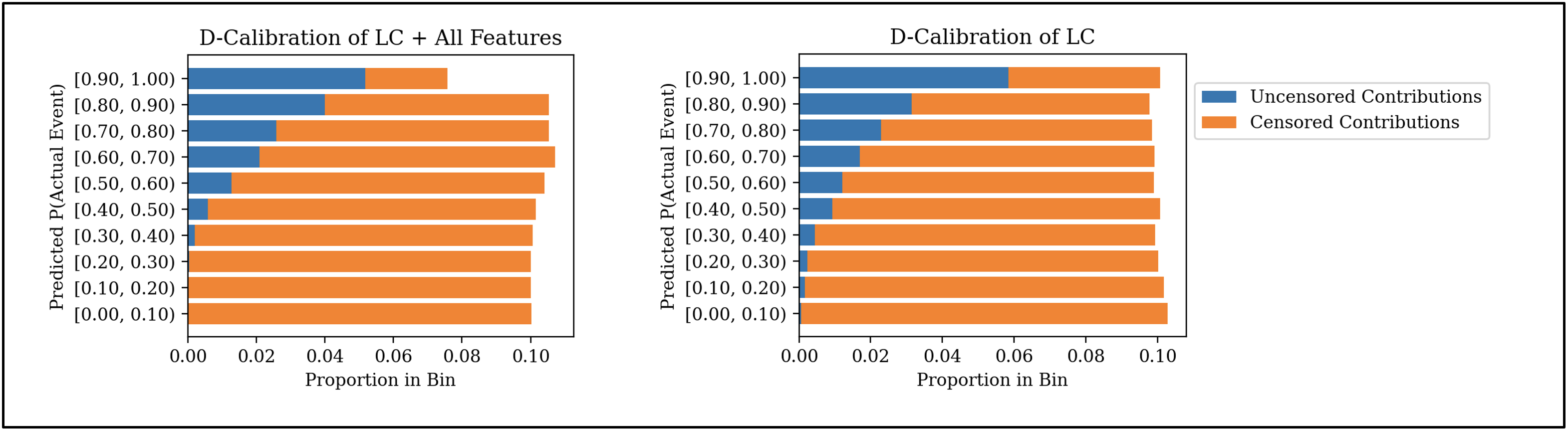}
    \caption{D-calibration histogram of LC + All Features model (left) and LC model (right), binned into ten groups. In a completely D-calibrated model,18 all of the horizontal bars should be at 0.10.}
    \label{fig:d_calibration}
\end{figure}

\begin{table}[]
    \centering
    \caption{Back-testing performance of All Features, LC + All Features, and LC only over 3-month seasons. Spring (SP) is March 20th until June 21st, followed by summer (SU) until September 22nd, followed by fall (F) until December 21st, followed by winter (W) until March 20th.}
    \label{tab:backtesting}
    \vspace{0.5em}
    \begin{tabular}{|l|cccccccc|}
\hline
\multirow{2}{*}{\begin{tabular}[c]{@{}l@{}}\textbf{All Features only,} \\ \textbf{trained up to:}\end{tabular}} & \multicolumn{8}{c|}{\textbf{Test C-index evaluated on:}}                                                                                                                                                                                              \\ \cline{2-9} 
                                                                                              & \multicolumn{1}{c|}{SP   2020} & \multicolumn{1}{c|}{SU   2020} & \multicolumn{1}{c|}{F   2020} & \multicolumn{1}{c|}{W   2020} & \multicolumn{1}{c|}{SP   2021} & \multicolumn{1}{c|}{SU   2021} & \multicolumn{1}{c|}{F   2021} & W   2021 \\ \hline
End   of spring 2020                                                                          & \multicolumn{1}{c|}{0.842}     & \multicolumn{1}{c|}{0.903}     & \multicolumn{1}{c|}{0.855}    & \multicolumn{1}{c|}{0.839}    & \multicolumn{1}{c|}{0.804}     & \multicolumn{1}{c|}{0.841}     & \multicolumn{1}{c|}{0.808}    & 0.845    \\ \hline
End   of summer 2020                                                                          & \multicolumn{1}{c|}{-}         & \multicolumn{1}{c|}{0.713}     & \multicolumn{1}{c|}{0.694}    & \multicolumn{1}{c|}{0.697}    & \multicolumn{1}{c|}{0.622}     & \multicolumn{1}{c|}{0.699}     & \multicolumn{1}{c|}{0.667}    & 0.711    \\ \hline
End   of fall 2020                                                                            & \multicolumn{1}{c|}{-}         & \multicolumn{1}{c|}{-}         & \multicolumn{1}{c|}{0.882}    & \multicolumn{1}{c|}{0.868}    & \multicolumn{1}{c|}{0.813}     & \multicolumn{1}{c|}{0.855}     & \multicolumn{1}{c|}{0.84}     & 0.907    \\ \hline
End   of winter 2020                                                                          & \multicolumn{1}{c|}{-}         & \multicolumn{1}{c|}{-}         & \multicolumn{1}{c|}{-}        & \multicolumn{1}{c|}{0.749}    & \multicolumn{1}{c|}{0.646}     & \multicolumn{1}{c|}{0.718}     & \multicolumn{1}{c|}{0.718}    & 0.735    \\ \hline
End   of spring 2021                                                                          & \multicolumn{1}{c|}{-}         & \multicolumn{1}{c|}{-}         & \multicolumn{1}{c|}{-}        & \multicolumn{1}{c|}{-}        & \multicolumn{1}{c|}{0.818}     & \multicolumn{1}{c|}{0.856}     & \multicolumn{1}{c|}{0.844}    & 0.908    \\ \hline
End   of summer 2021                                                                          & \multicolumn{1}{c|}{-}         & \multicolumn{1}{c|}{-}         & \multicolumn{1}{c|}{-}        & \multicolumn{1}{c|}{-}        & \multicolumn{1}{c|}{-}         & \multicolumn{1}{c|}{0.859}     & \multicolumn{1}{c|}{0.847}    & 0.91     \\ \hline
End   of fall 2021                                                                            & \multicolumn{1}{c|}{-}         & \multicolumn{1}{c|}{-}         & \multicolumn{1}{c|}{-}        & \multicolumn{1}{c|}{-}        & \multicolumn{1}{c|}{-}         & \multicolumn{1}{c|}{-}         & \multicolumn{1}{c|}{0.850}    & 0.911    \\ \hline
1/12/2022   (study end)                                                                       & \multicolumn{1}{c|}{-}         & \multicolumn{1}{c|}{-}         & \multicolumn{1}{c|}{-}        & \multicolumn{1}{c|}{-}        & \multicolumn{1}{c|}{-}         & \multicolumn{1}{c|}{-}         & \multicolumn{1}{c|}{-}        & 0.912    \\ \hline
\end{tabular}
\newline
\vspace*{1 cm}
\newline
\begin{tabular}{|l|cccccccc|}
\hline
\multirow{2}{*}{\textbf{\begin{tabular}[c]{@{}l@{}}All Features + LCs, \\ trained up to:\end{tabular}}} & \multicolumn{8}{c|}{\textbf{Test C-index evaluated on:}}                                                                                                                                                                                     \\ \cline{2-9} 
                                                                                                        & \multicolumn{1}{c|}{SP   2020} & \multicolumn{1}{c|}{SU   2020} & \multicolumn{1}{c|}{F   2020} & \multicolumn{1}{c|}{W   2020} & \multicolumn{1}{c|}{SP   2021} & \multicolumn{1}{c|}{SU   2021} & \multicolumn{1}{c|}{F   2021} & W   2021 \\ \hline
End   of spring 2020                                                                                    & \multicolumn{1}{c|}{0.791}     & \multicolumn{1}{c|}{0.840}     & \multicolumn{1}{c|}{0.852}    & \multicolumn{1}{c|}{0.837}    & \multicolumn{1}{c|}{0.774}     & \multicolumn{1}{c|}{0.814}     & \multicolumn{1}{c|}{0.805}    & 0.876    \\ \hline
End   of summer 2020                                                                                    & \multicolumn{1}{c|}{-}         & \multicolumn{1}{c|}{0.847}     & \multicolumn{1}{c|}{0.852}    & \multicolumn{1}{c|}{0.845}    & \multicolumn{1}{c|}{0.775}     & \multicolumn{1}{c|}{0.806}     & \multicolumn{1}{c|}{0.818}    & 0.891    \\ \hline
End   of fall 2020                                                                                      & \multicolumn{1}{c|}{-}         & \multicolumn{1}{c|}{-}         & \multicolumn{1}{c|}{0.852}    & \multicolumn{1}{c|}{0.845}    & \multicolumn{1}{c|}{0.781}     & \multicolumn{1}{c|}{0.812}     & \multicolumn{1}{c|}{0.822}    & 0.896    \\ \hline
End   of winter 2020                                                                                    & \multicolumn{1}{c|}{-}         & \multicolumn{1}{c|}{-}         & \multicolumn{1}{c|}{-}        & \multicolumn{1}{c|}{0.846}    & \multicolumn{1}{c|}{0.778}     & \multicolumn{1}{c|}{0.811}     & \multicolumn{1}{c|}{0.823}    & 0.896    \\ \hline
End   of spring 2021                                                                                    & \multicolumn{1}{c|}{-}         & \multicolumn{1}{c|}{-}         & \multicolumn{1}{c|}{-}        & \multicolumn{1}{c|}{-}        & \multicolumn{1}{c|}{0.778}     & \multicolumn{1}{c|}{0.813}     & \multicolumn{1}{c|}{0.827}    & 0.899    \\ \hline
End   of summer 2021                                                                                    & \multicolumn{1}{c|}{-}         & \multicolumn{1}{c|}{-}         & \multicolumn{1}{c|}{-}        & \multicolumn{1}{c|}{-}        & \multicolumn{1}{c|}{-}         & \multicolumn{1}{c|}{0.812}     & \multicolumn{1}{c|}{0.826}    & 0.899    \\ \hline
End   of fall 2021                                                                                      & \multicolumn{1}{c|}{-}         & \multicolumn{1}{c|}{-}         & \multicolumn{1}{c|}{-}        & \multicolumn{1}{c|}{-}        & \multicolumn{1}{c|}{-}         & \multicolumn{1}{c|}{-}         & \multicolumn{1}{c|}{0.827}    & 0.900    \\ \hline
1/12/2022   (study end)                                                                                 & \multicolumn{1}{c|}{-}         & \multicolumn{1}{c|}{-}         & \multicolumn{1}{c|}{-}        & \multicolumn{1}{c|}{-}        & \multicolumn{1}{c|}{-}         & \multicolumn{1}{c|}{-}         & \multicolumn{1}{c|}{-}        & 0.900    \\ \hline
\end{tabular}
\newline
\vspace*{1 cm}
\newline
\begin{tabular}{|l|cccccccc|}
\hline
\multirow{2}{*}{\textbf{\begin{tabular}[c]{@{}l@{}}LCs only, \\ trained up to:\end{tabular}}} & \multicolumn{8}{c|}{\textbf{Test C-index evaluated on:}}                                                                                                                                                                                     \\ \cline{2-9} 
                                                                                              & \multicolumn{1}{c|}{SP   2020} & \multicolumn{1}{c|}{SU   2020} & \multicolumn{1}{c|}{F   2020} & \multicolumn{1}{c|}{W   2020} & \multicolumn{1}{c|}{SP   2021} & \multicolumn{1}{c|}{SU   2021} & \multicolumn{1}{c|}{F   2021} & W   2021 \\ \hline
End   of spring 2020                                                                          & \multicolumn{1}{c|}{0.797}     & \multicolumn{1}{c|}{0.863}     & \multicolumn{1}{c|}{0.869}    & \multicolumn{1}{c|}{0.850}    & \multicolumn{1}{c|}{0.795}     & \multicolumn{1}{c|}{0.833}     & \multicolumn{1}{c|}{0.83}     & 0.904    \\ \hline
End   of summer 2020                                                                          & \multicolumn{1}{c|}{-}         & \multicolumn{1}{c|}{0.857}     & \multicolumn{1}{c|}{0.863}    & \multicolumn{1}{c|}{0.842}    & \multicolumn{1}{c|}{0.785}     & \multicolumn{1}{c|}{0.819}     & \multicolumn{1}{c|}{0.822}    & 0.904    \\ \hline
End   of fall 2020                                                                            & \multicolumn{1}{c|}{-}         & \multicolumn{1}{c|}{-}         & \multicolumn{1}{c|}{0.867}    & \multicolumn{1}{c|}{0.850}    & \multicolumn{1}{c|}{0.800}     & \multicolumn{1}{c|}{0.828}     & \multicolumn{1}{c|}{0.831}    & 0.909    \\ \hline
End   of winter 2020                                                                          & \multicolumn{1}{c|}{-}         & \multicolumn{1}{c|}{-}         & \multicolumn{1}{c|}{-}        & \multicolumn{1}{c|}{0.854}    & \multicolumn{1}{c|}{0.801}     & \multicolumn{1}{c|}{0.826}     & \multicolumn{1}{c|}{0.834}    & 0.911    \\ \hline
End   of spring 2021                                                                          & \multicolumn{1}{c|}{-}         & \multicolumn{1}{c|}{-}         & \multicolumn{1}{c|}{-}        & \multicolumn{1}{c|}{-}        & \multicolumn{1}{c|}{0.803}     & \multicolumn{1}{c|}{0.828}     & \multicolumn{1}{c|}{0.834}    & 0.911    \\ \hline
End   of summer 2021                                                                          & \multicolumn{1}{c|}{-}         & \multicolumn{1}{c|}{-}         & \multicolumn{1}{c|}{-}        & \multicolumn{1}{c|}{-}        & \multicolumn{1}{c|}{-}         & \multicolumn{1}{c|}{0.829}     & \multicolumn{1}{c|}{0.833}    & 0.910    \\ \hline
End   of fall 2021                                                                            & \multicolumn{1}{c|}{-}         & \multicolumn{1}{c|}{-}         & \multicolumn{1}{c|}{-}        & \multicolumn{1}{c|}{-}        & \multicolumn{1}{c|}{-}         & \multicolumn{1}{c|}{-}         & \multicolumn{1}{c|}{0.837}    & 0.914    \\ \hline
1/12/2022   (study end)                                                                       & \multicolumn{1}{c|}{-}         & \multicolumn{1}{c|}{-}         & \multicolumn{1}{c|}{-}        & \multicolumn{1}{c|}{-}        & \multicolumn{1}{c|}{-}         & \multicolumn{1}{c|}{-}         & \multicolumn{1}{c|}{-}        & 0.913    \\ \hline
\end{tabular}
\end{table}

\section{Discussion}

\paragraph{Learned Concept Classifiers.}  The strongest coefficients for each concept classifier are often not features that immediately come to mind, but nevertheless match clinical intuition. For the old age concept, we observe that a high PF flu vaccine has a large coefficient (3.31), and is a vaccine only given to patients 65 and older. The outpatient shortness of breath symptom (3.23) is the second highest coefficient for the inpatient concept, possibly indicating that outpatients with this symptom are at high risk of becoming an inpatient. The shortness of breath concept depends on dexamethasone (0.75) which relieves inflammation, and albuterol sulfate (0.67), prescribed for lung conditions. For the obesity concept, sleep apnea, which is often caused by excess weight, has the largest coefficient (0.96).

Note, however, that the learned concepts may not be a perfect representation of the underlying concept. None of the concept classifiers perfectly recover the originally known positives, with recall ranging from 0.381 to 0.974 (Table \ref{tab:num_pu_pos}). This could be due to insufficient signal in the remaining covariates or underfitting due to the simplicity of the logistic regression model class. Additionally, some concepts learn substantially more positives than were available in the original data. For example, obesity originally has 433 positives in the data but the concept classifier marks 2,157 patients as having obesity with probability greater than 0.5. It is difficult to verify the faithfulness of the concept classifiers to the true concepts without manual review, but it is possible that the learned concepts may mark patients as “obesity-like” based on their other covariates rather than learning whether they truly have obesity. Additionally, while learned concepts are amenable to interpretation through the coefficients of the concept classifier, they still require domain expertise to manually define positive anchor variables. Finally, the conditional independence assumptions of the selected anchors may not hold in practice, and these assumptions are difficult to verify.

\paragraph{Learned Survival Models.}  The coefficients for the survival models trained on LCs, All Features, and LC + All Features are mostly consistent with clinical intuition. Inpatients are more likely to experience adverse outcomes than those not hospitalized, old age is well-documented to be associated with higher COVID-19 death rates,\cite{1docherty2020features,7liang2020development,8galloway2020clinical} shortness of breath indicates respiratory involvement, and medications such as dexamethasone, acetaminophen, and intravenous saline are given to hospitalized patients. Higher BUN indicates worse liver and kidney function, and COVID-19 vaccines are designed to protect against severe COVID-19. While it is surprising that some of the learned concepts for fever and cough symptoms have negative HRs, upon inspection we find that these are most reliably recorded for outpatients and may encode some additional information about outpatient status. From exploring the interactive visualization, some of the features selected by the All Features model are used by LCs in the LC + All Features model, possibly indicating that they serve as proxies for higher level concepts. For example, the saline IV bolus, present only in the All Features model, is used in the inpatient concept classifier with a positive coefficient. We also note that the coefficients are likely shrunken towards zero due to the Lasso penalty, and the non-informative or independent censoring assumption of the Cox proportional hazards model may not hold since censoring occurs upon discharge. 

Our Lasso-Cox models all outperform the baselines (Covichem, Galloway count, Galloway reweighted) in terms of aggregate, inpatient, and outpatient concordance (Table \ref{tab:model_perf}). As measured by aggregate concordance, we observe that the learned concepts provide a boost in performance over the raw positive anchors (C-index 0.858 vs. 0.844). This boost in performance places the LC model approximately halfway between the performance of the raw positive anchors and the LC + All Features and All Features models, which both achieve a C-index of 0.872. For LC + All Features and All Features models, the C-index on the outpatient subpopulation (0.879 and 0.880) is higher than that on the entire cohort, whereas the C-index on the inpatient subpopulation (0.715 and 0.717) is lower. For all remaining models, the performance on the inpatient and outpatient subpopulations is lower than in aggregate, possibly indicating that it is easy to order the relative risks of inpatients versus outpatients. The LC model appears slightly better calibrated than the LC + All Features model, but when used to stratify patients into high, medium, and low-risk strata, both models yield groups with clear separation between their survival curves. Finally, while there is some loss of discriminative performance going from the All Features models to the LC only model, when tested under the back-testing framework this gap seems to close quickly on subsequent time periods and the LC model even eventually surpasses the performance of the All Features model. Thus, models with LCs might be more resilient over time than models learned only from All Features. If the set of important high-level concepts themselves change over time, however, new concepts may need to be learned accordingly.

\paragraph{Future Work.}  In the future, we plan to integrate our models with the healthcare system, and to continue to monitor the performance of the models over time. It will be important to further study how high-level clinical concepts perform across different settings, and to quantify the extent to which classifiers learned on top of these concepts might be transferable across hospitals. As COVID-19 continues to evolve over time, we will also investigate whether new concepts become relevant for prediction.

\newpage
\bibliographystyle{abbrvnat}
\bibliography{refs}
\end{document}